\def\year{2021}\relax
\documentclass[letterpaper]{article} % DO NOT CHANGE THIS
\usepackage{aaai21}  % DO NOT CHANGE THIS
\usepackage{times}  % DO NOT CHANGE THIS
\usepackage{helvet} % DO NOT CHANGE THIS
\usepackage{courier}  % DO NOT CHANGE THIS
\usepackage[hyphens]{url}  % DO NOT CHANGE THIS
\usepackage{graphicx} % DO NOT CHANGE THIS
\usepackage{natbib}  % DO NOT CHANGE THIS AND DO NOT ADD ANY OPTIONS TO IT
\usepackage{caption} % DO NOT CHANGE THIS AND DO NOT ADD ANY OPTIONS TO IT
\newcommand{\etal}{\emph{et al. }}
\newcommand{\eg}{\emph{e.g. }}
\newcommand{\ie}{\emph{i.e. }} 
\usepackage{times}
\usepackage{epsfig}
\usepackage{graphicx}
\usepackage{multirow}
\usepackage{amsmath}
\usepackage{amssymb}
\usepackage{boldline}
\usepackage[switch]{lineno}
\usepackage{soul}
\newcommand{\R}{\mathbb{R}}
\usepackage[ruled]{algorithm2e}

\title{CompFeat: Comprehensive Feature Aggregation for Video Instance Segmentation}
\author{Yang Fu$^{1}$,  Linjie Yang$^{2}$, Ding Liu$^{2}$, Thomas S. Huang$^{1}$, Humphrey Shi$^{1,3}$ \\ 
{\small $^1$University of Illinois at Urbana-Champaign},
{\small $^2$ByteDance Inc.}, {\small $^3$University of Oregon}\\ }
\begin{document}
\maketitle
% \linenumbers
%%%%%%%%% ABSTRACT
% \input{TEX/title.tex}
\begin{abstract}
Video instance segmentation is a complex task in which we need to detect, segment, and track each object for any given video. Previous approaches only utilize single-frame features for the detection, segmentation, and tracking of objects and they suffer in the video scenario due to several distinct challenges such as motion blur and drastic appearance change. 
To eliminate ambiguities introduced by only using single-frame features, we propose a novel comprehensive feature aggregation approach (\textbf{CompFeat}) to refine features at both frame-level and object-level with temporal and spatial context information. 
The aggregation process is carefully designed with a new attention mechanism which significantly increases the discriminative power of the learned features.
We further improve the tracking capability of our model through a siamese design by incorporating both feature similarities and spatial similarities. Experiments conducted on the YouTube-VIS dataset validate the effectiveness of proposed CompFeat.
% Our proposed CompFeat achieves the state-of-the-art results on the challenging YouTube-VIS dataset.
Our code will be available at \url{https://github.com/SHI-Labs/CompFeat-for-Video-Instance-Segmentation}.

\end{abstract}

%%%%%%%%% BODY TEXT
\section{Introduction}
Video instance segmentation (VIS) is a joint task of detection, segmentation and tracking of object instances in videos~\cite{yang2019video}. Different from instance segmentation in image domain~\cite{hariharan2014simultaneous}, video instance segmentation not only requires to segment object masks on individual frames, but also to track the identities of objects across different frames. Also, unlike semi-supervised video object segmentation~\cite{voigtlaender2019feelvos, voigtlaender2017online, xu2018youtube, wug2018fast, oh2019video, xu2019spatiotemporal}, video instance segmentation does not require a ground truth mask in the first frame and all objects appear in the video should be processed. It has essential applications in many video-based tasks, including video editing, autonomous driving and augmented reality.

Video instance segmentation has several distinct challenges. For example, if an object is recognized as a wrong category in one frame of the video, tracking of this object will be extremely hard due to inconsistency of object categories. When there are multiple similar objects, finding the correspondences of them across the video is also challenging.
VIS is an important but underexplored task. The pioneering work for VIS is MaskTrack-RCNN~\cite{yang2019video}, which is built upon Mask-RCNN~\cite{he2017mask}, a state-of-the-art method for image instance segmentation. 
A new tracking branch is tailored and added in order to track object instances.
However, MaskTrack-RCNN relies on only single frame object features and neglects critical temporal information, \ie temporal consistency of an object, motion pattern of different objects, etc, all of which can provide abundant information for category recognition, object detection and mask segmentation across video frames. And lots of recent video understanding work~\cite{wang2018non, wu2019long} focused on how to utilize the temporal information.
In addition, the proposed tracking head in MaskTrack-RCNN is preliminary and ignores the spatial layout of objects with simple object features, 
which has been proven crucial by modern object tracking algorithms~\cite{zhang2019deeper,Zhu_2018_ECCV,li2019siamrpn++} to improve the tracking of video instances.
%from fully connected layers. 
%
%The modern video tracking algorithms~\cite{zhang2019deeper,Zhu_2018_ECCV,li2019siamrpn++} prove that the spatial information of target objects is crucial and can be potentially leveraged to help improve the tracking of video instances.

\begin{figure}[t]
	\centering
	\includegraphics[width=0.45\textwidth]{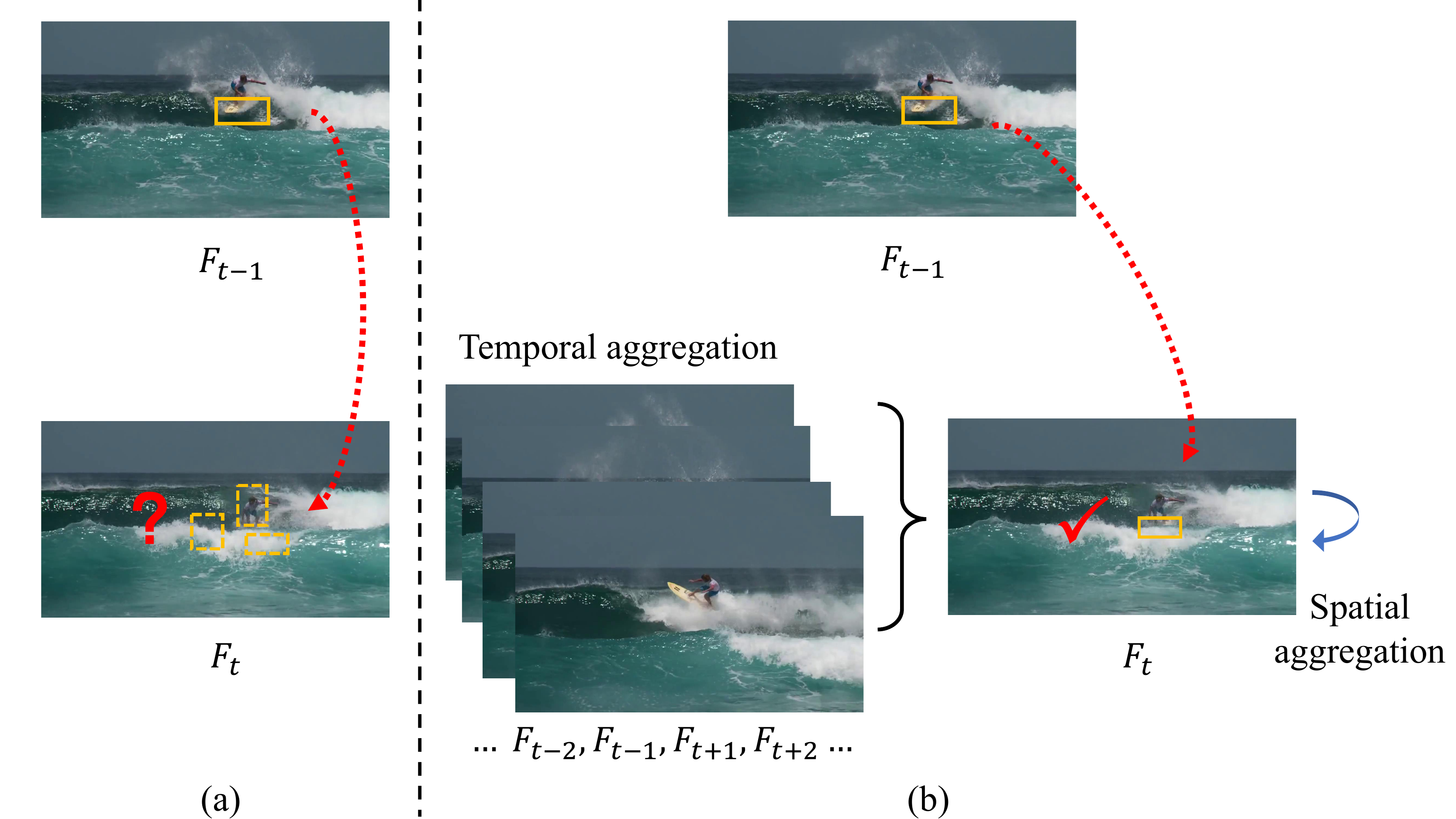}
% 	\vspace{3mm}
	\caption{An illustration of the proposed comprehensive feature aggregation (CompFeat) method. $F_t$ denotes the video frame at time $t$. (a) Previous video instance segmentation method without feature aggregation. (b) Our proposed comprehensive feature aggregation approach for video instance segmentation}
	\label{fig:illus}
% 	\vspace{-3mm}
\end{figure}

In order to utilize the abundant information in videos and to harvest the benefits of modern object tracking approaches, we propose a comprehensive feature aggregation approach for video instance segmentation, termed \textbf{CompFeat}. The main idea of CompFeat is illustrated in Fig~\ref{fig:illus}.  
As shown in Fig~\ref{fig:illus}(a), the key object is not detected and fails to be tracked due to using only the unclear visual cues of a single frame, while other frames in the same video contain helpful information for locating and tracking the correct object. 
Hence, we propose a dual attention module with both temporal attention and spatial attention to aggregate contextual information from neighboring frames and other positions in the current frame as described in Fig~\ref{fig:illus}(b). 
We also enhance the features of detected objects by extending the dual attention module to object level, which substantially improves the discriminative power of the object features, enabling more accuracy object detection and segmentation. 
In addition, we introduce a novel correlation-based tracking module to improve instance tracking across different frames. Instead of using a holistic similarity between a pair of detected object and reference object to determine object correspondence, our correlation-based module not only employs depth-wise correlation between an object pair to generate a matching score with spatial awareness, but also computes a correlation map between a reference object and the current frame to better localize the target object similar to Siamese object tracking.

To summarize, the main contributions of this work are threefold as follows:
\begin{itemize}
\item We propose a comprehensive feature aggregation approach for video instance segmentation, including temporal and spatial attention modules on both frame-level and object-level features.
\item We introduce a correlation-based tracking module to track instances across frames, which predicts cross-correlation maps in both object-to-object and object-to-frame manners to produce multiple similarity cues for object tracking.
\item We conduct extensive experiments and ablation study on YouTube-VIS~\cite{yang2019video} to demonstrate the effectiveness of our proposed framework and each of the individual components.
\end{itemize}

% The rest of our paper is organized as follows. In Section~\ref{related} we briefly introduce the recent achievement on video instance segmentation and some related task. Then, in Section~\ref{method}, we state and discuss how our proposed ComFeat method works in details. Finally, several experimental results and ablation studies are presented in Section~\ref{exp}.
%------------------------------------------------------------------------
\section{Related Work}~\label{related}
In this section we review  video instance segmentation and several closely-related tasks such as video object detection and video object tracking.

{\bf Video Object Detection.} Video object detection aims to detect all objects in videos such as shown in the ImageNet VID challenge~\cite{russakovsky2015imagenet,han2016seq}. Feature aggregation is widely used in video detection~\cite{zhu2017flow,feichtenhofer2017detect,chen2018optimizing,liu2019looking}. For instance, Zhu~\etal proposed to aggregate features from nearby frames to enhance the feature quality of an input frame. However, its speed is pretty slow due to the dense detection and optical flow estimation. In~\cite{chen2018optimizing}, Chen~\etal proposed to use a scale-time lattice to generate detection on sparse key frames and designed a temporal propagation approach for detection in an effective way. Inspired by these work, we propose to improve the feature quality for video instance segmentation via feature aggregation using attention mechanism.

{\bf Video Object Tracking}. Video object tracking can be viewed as a sub-task of video instance segmentation, which has two scenarios: detection-based tracking and detection-free tracking. In detection-free tracking, given the ground truth location of the target object in the first frame, algorithm is required to track the target object through the whole video. Recently, the Siamese network based trackers have received significant attentions due to their well-balanced accuracy and efficiency~\cite{li2019siamrpn++, zhang2019deeper, wang2018learning, valmadre2017end}. In particular, these trackers attempt to produce a similarity map from cross-correlation of the two feature branches, one for the target object and the other for the search region, where the similarity map embeds more semantic meanings. On the other hand, detection-based tracking~\cite{sadeghian2017tracking, son2017multi,shi2018geometry} simultaneously detect and track multiple video objects, which is more similar to the setting of video instance segmentation. In our proposed CompFeat, we borrow ideas from both detection-based tracking and detection-free tracking.

{\bf Video Instance Segmentation}. MaskTrack-RCNN~\cite{yang2019video} is the first attempt to address the video instance segmentation problem. It proposes a large-scale video dateset named YouTube-VIS for benchmarking video instance segmentation algorithms. Several methods in the Large-Scale Video Object Segmentation Challenge achieve impressive results but they utilize large quantity of external data and complex algorithm pipelines~\cite{wang2019empirical, dong2019temporal, luiten2019video}. A closely related work is multi-object tracking and segmentation~\cite{voigtlaender2019mots} which is proposed to evaluate multi-object tracking along with instance segmentation. However, because of its limited data scale and few object categories, we do not compare with it in this paper. MaskTrack-RCNN only uses image features but not temporal information of video sequences. We extend this work with a more sophisticated comprehensive feature aggregation approach which greatly boosts the performance on video instance segmentation.

\begin{figure*}[t]
	\centering
	\includegraphics[width=0.65\textwidth]{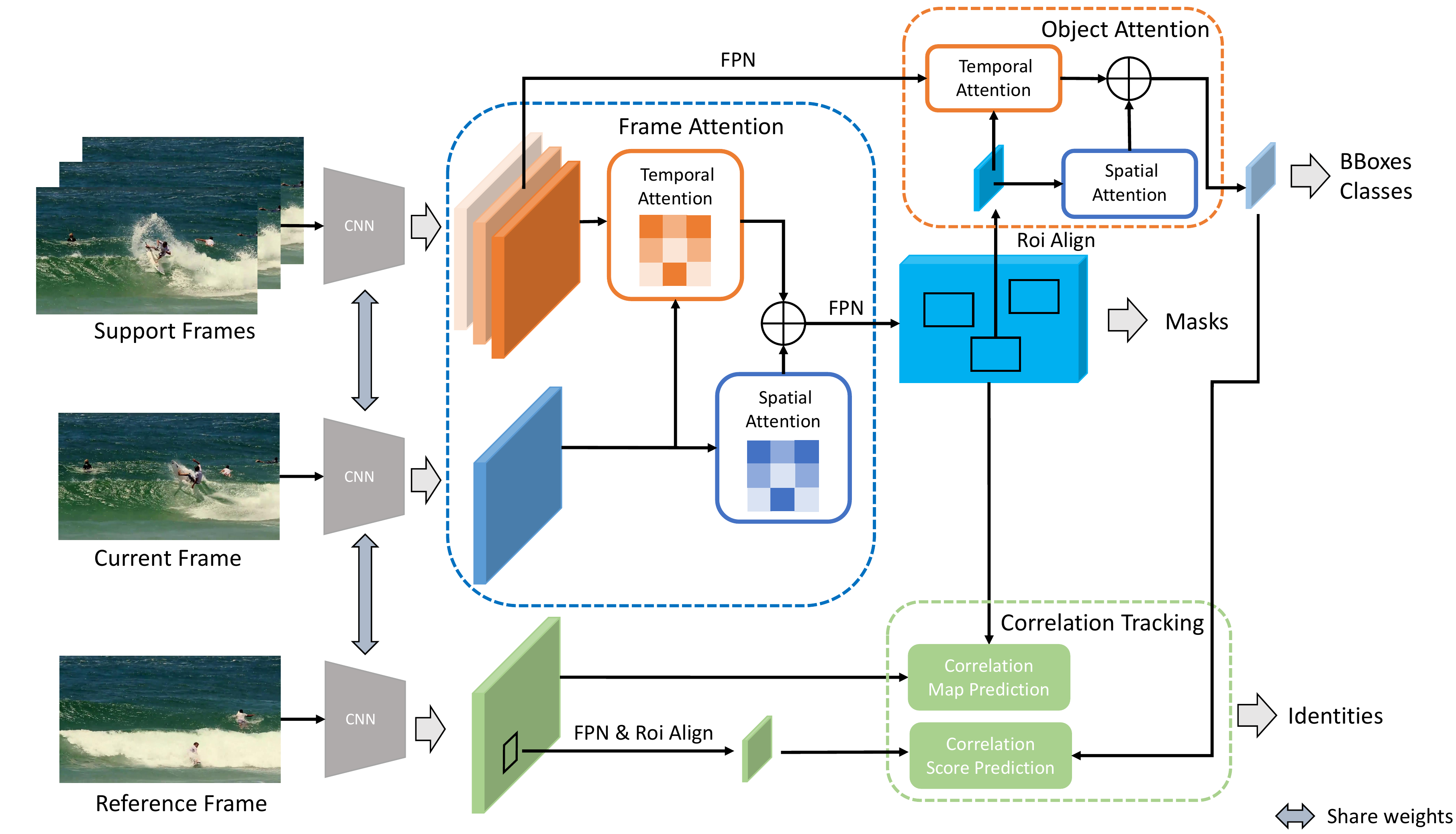}
% 	\vspace{3mm}
	\caption{An overview of our proposed CompFeat approach for video instance segmentation. CompFeat consists of three major components: (a) A frame level attention module incorporating both temporal and spatial attention modules. (b) An object level attention module which has a similar structure as frame level attention. (c) A correlation-based tracking module to predict the correlation score and correlation map simultaneously.}
	\label{fig:frame}
% 	\vspace{-3mm}
\end{figure*}
%------------------------------------------------------------------------
\section{Proposed Method}\label{method}
In our framework, the video frames are sequentially processed. During the video processing, we randomly select one frame as the current frame and sample other several frames as support frames which are used for temporal feature aggregation. Meanwhile, a correlation-based tracking module is proposed to track object identities across frames with comprehensive cues. The overview of the proposed CompFeat framework is shown in Fig~\ref{fig:frame}. The current frame and all support frames are first fed into ResNet50~\cite{he2016deep} for feature extraction. Then, our proposed {\bf temporal attention module} takes the features of the current frame and support frames as inputs for feature aggregation over different frames. Meanwhile, the features of the current frame are processed by a {\bf spatial attention module} for global context feature aggregation on a single frame. The similar process is performed on object level with same model structures. In addition, we enhance the tracking branch of the network via a {\bf correlation-based tracking module} for a more accurate object tracking. The correlation-based tracking module combines cross-correlation between a pair of reference object and newly detected object and correlation between the reference object and the current frame (search region). Finally, we integrate the three proposed modules into a complete framework to perform three different tasks, object detection, mask segmentation and object tracking simultaneously. Next we describe each proposed component in details.

\subsection{Temporal Attention Module}
\begin{figure}[t]
	\centering
	\includegraphics[width=0.4\textwidth]{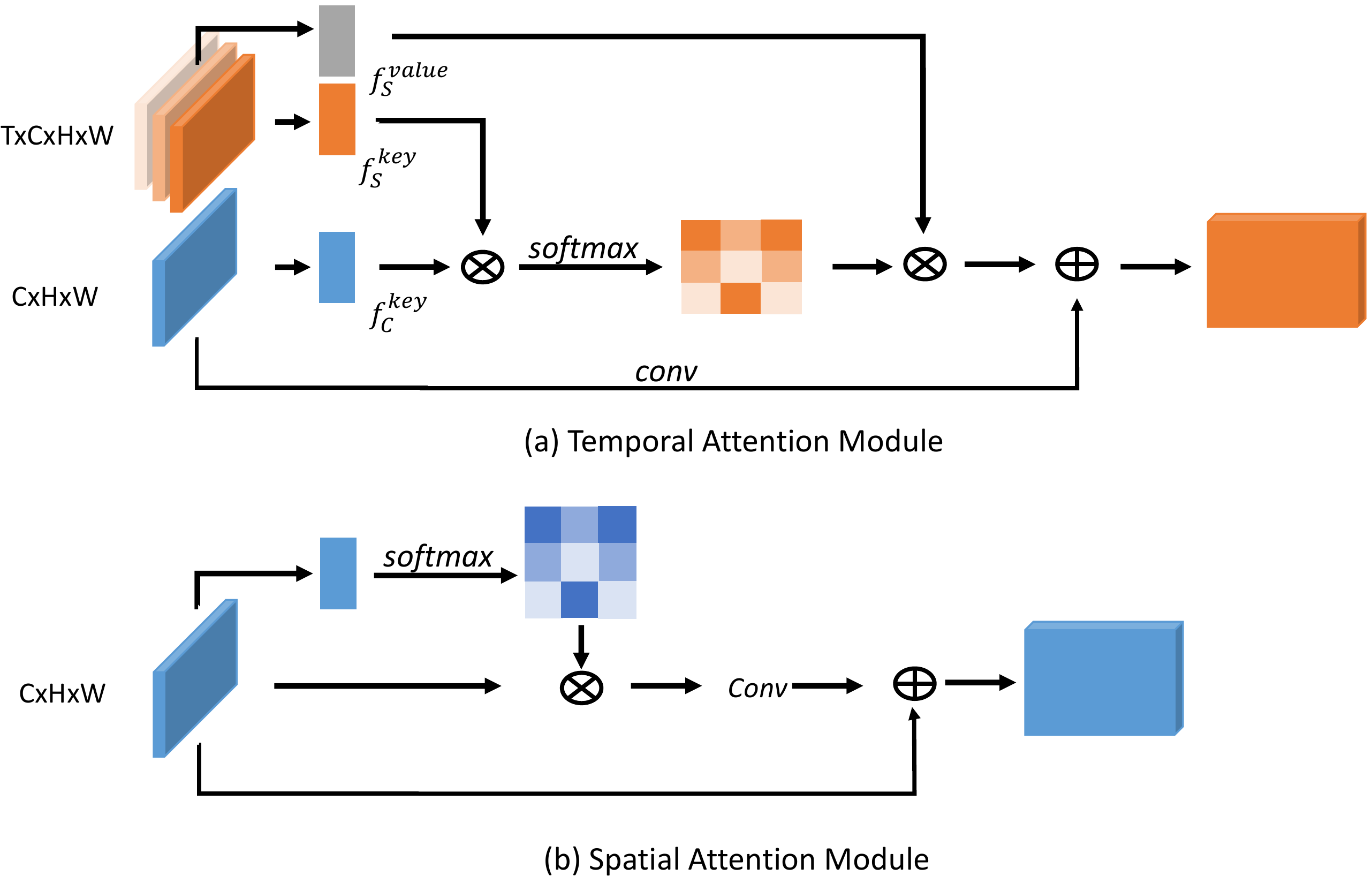}
% 	\vspace{3mm}
	\caption{An illustration of (a) temporal attention module and (b) spatial attention module. Best viewed in color.}
	\label{fig:att}
\end{figure}

Inspired by the non-local networks~\cite{wang2018non}, we propose a novel temporal attention module to refine the features of the current frame via aggregating information from other support frames. Different from the original non-local block, which aims to model the long-range dependencies by attention mechanism, our proposed temporal attention module focuses more on embedding information from other frames and use them to refine features of the current frame by cross-attention mechanism. Specifically, the temporal attention module has three steps: embedding of current frame embedding, embedding of support frames and features aggregation as described in Fig~\ref{fig:att}(a).

{\bf Embedding of current frame}. Given the features of the current frame $f_{C} \in \R^{C\times H \times W}$, where $H, W, C$ are the height, width, and the feature dimension of output feature map from the backbone network. We first feed it into a convolution layer to generate a feature map $f_{C}^{key}$, where $f_{C}^{key} \in \R^{\frac{C}{4} \times H \times W}$, and a non-linearity activation is applied. The $f_{C}^{key}$ can store the key features of the current frame including which and where objects may exist. Therefore, the feature map $f_{C}^{key}$ is learnt to encode the key information of visual semantics in the current frame, \ie object categories, object locations and masks.

{\bf Embedding of support frames}. Given a stack of feature maps obtained from support frames $\{f_{S_{t}} \in \R^{C\times H \times W}; t=1:T\}$ ($T$ is the number of the support frames), each feature map $f_{S_t}$ is first encoded into a pair of feature maps $f_{S_{t}}^{key}$ and $f_{S_{t}}^{value}$ by two parallel convolution layers, where $f_{S_{t}}^{key}, f_{S_{t}}^{value} \in \R^{\frac{C}{4} \times H \times W}$.  Non-linear activation is applied as well. If there are more than one support frames ($T>1$), we concatenate features of different frames along temporal dimension and obtain $f_S^{key}, f_S^{value}  \in \R ^{T \times \frac{C}{4} \times H \times W}$. $f_{S}^{key}$ contains the information of key features of support frames. The similarities between $f_{C}^{key}$ and $f_{S}^{key}$ shows when-and-where the features of support frames are suitable to be aggregated for the current frame. In addition, the feature map $f_{S}^{value}$ is learnt to represent the context information of all support frames.

{\bf Features aggregation}. In the feature aggregation step, we first compute the attention weights by similarities between all pixel in feature map $f_{C}^{key}$ and $f_{S}^{key}$. The similarities are computed as the correlation of every spatial location in the feature map $f_{C}^{key}$ and every spatial-temporal location in feature map $f_{S}^{key}$. Then the attention weights are used to aggregate features from $f_{S}^{value}$ to obtain context features from support frames. In addition, we perform a feature transformation by a $1 \times 1$ convolution layer. The whole process of feature aggregation for every position of the current frame can be summarized as the following equation,
\begin{equation}
F_{TA} \Leftrightarrow 
\label{eq1}
\left\{
\begin{aligned}
& X = f_S^{key} \odot f_{C}^{key} \\
& f_A(:, j) = F(f_S^{value} \odot \frac{exp(X(:, j))}{\sum_{i=1}^{N_p} exp(X(:, i))}) 
\end{aligned}
\right. 
\end{equation}
% \begin{equation}
% \begin{aligned}
% \tilde{f_C} = & f_A + f_C
% \end{aligned}
% \end{equation}
where $X$ is the similarity matrix between the current frame and support frames with size of $HWT \times HW$, $i$, $j$ are the indices of every position in the similarity matrix and the feature map, $N_p$ is the total number of positions in the feature map, $\odot$ is dot product, $F$ is a transformation function with non-linear activation, $f_A$ is the aggregated feature map after the transformation. Note that, all feature maps in above equation may be processed by some necessary reshaping or permutation operations. Finally, the refined feature map $\tilde{f_C}$ is obtained by summing up the aggregated feature map $f_A$ and the feature map of the current frame $f_C$.
 
After aggregation, we can obtain a feature map $\tilde{f_C}$, which not only preserves some information key visual semantics of current frame, but also extracts useful contextual information existing in other frames within the same video. 

\subsection{Spatial Attention Module}
Besides aggregating temporal contextual information from support frames, we also propose a spatial attention module to model the spatial context of the current frame. The proposed spatial attention module is based on the non-local block~\cite{wang2018non} with several 
modifications. As shown in Fig~\ref{fig:att}(b), instead of computing the attention maps of all spatial positions for each channel, we simplify it by sharing a single attention weight for all spatial positions in each channel, which is known as the channel attention. As reported in~\cite{cao2019gcnet}, this simplified channel attention module can achieve very similar performance compared to original non-local block. Also, since it doesn't need to compute a specific attention weight for every spatial position, it can be more efficient. Then, we use two more convolution layers to transform the features. Finally, the channel attention is added back to the feature map of the current frame for global context feature aggregation. 
% \lj{is it more efficient then the original non-local block? why do you use this one?} 

\subsection{Combining Two Attention Modules}
Furthermore, we integrate the two attention modules into a dual attention module. The dual attention model takes the features of the current frame and support frames as inputs. The features of the current frame $f_C$ is first processed by spatial attention model to obtain an attention map with context features of itself. And features of current frames $f_C$ and features of support frames $f_S$ are fed into the temporal attention module to generate another attention map with context information of support frames. Then, we aggregate the features of the current frame with two attention maps by adding them to the original feature map $f_C$ as follows,  
\begin{equation}
\begin{aligned}\label{eq2}
& f_{agg} = F_{TA}(f_C, f_S) + F_{SA}(f_C) + f_C \\
\end{aligned}
\end{equation}
where $F_{TA}, F_{SA}$ represent proposed temporal attention module and spatial attention module respectively.

\subsection{Attention Module on Object Level}\label{sec3.4}
Attention modules described above are all processed on frame level features, which means they aggregate the context information for the whole feature map from the current frame and support frames. However, for instance segmentation with a two-stage framework, aggregating context information for each object proposal is also critical. Object proposals act as valuable candidates for the final object predictions and they encode more focused features for the individual objects. Proper feature aggregation onto these object proposals can elucidate confusing feature representations and improve recognition accuracy. We attempt to extend the two proposed attention modules to object level by applying similar operations to the object features produced by ROI Align~\cite{he2017mask}. Since the proposed attention modules can be applied to feature maps with arbitrary size, we adopt the two proposed attention modules in a similar fashion. 
We denote the features of detected object proposals as $f_{C_p}$. $f_{C_p} \in \R^{P\times C \times h \times w}$, where $P$ is the number of object proposals, $C$ is the channel dimension, and $h$, $w$ are height and width of the proposal. The features of each proposal are fed into a dual attention module along with feature maps of support frames. Since the size of the features of proposals is much smaller than the whole feature map, aggregating features on object level can be very efficient. In the next section, we show the performance gain of adding object-level attention modules is comparable or even better than adding attention module on frame level. 

The proposed attention module is inspired by some early work of video object detection and video object segmentation~\cite{wu2019long,oh2019video}, where the non-local blocks are used for extracting self-attention features. However, the motivation of proposed attention module is different as it is designed to aggregate the features from other frames (support frames) to current frame through feature matching, but not to itself by self-attention mechanisms. In addition, we extend the proposed attention module to object level, which largely reduces the computational complexity while improves the performance as frame-level attention module. Furthermore, frame-level and object-level module can be integrated into a single framework for a better performance. To our best knowledge, there are few work on enhancing object level feature by other frames. And the experiments in Sec.~\ref{exp} show the effectiveness of our proposed attention module.

By performing proposed dual attention module on both frame level and object level, we achieve feature aggregation on both spatial and temporal dimension, and in both local and global granularities, which is a comprehensive approach to aggregate and enhance intermediate features in a video instance segmentation framework.

\subsection{Correlation-based Tracking Module}

\begin{figure}[t]
	\centering
	\includegraphics[width=0.4\textwidth]{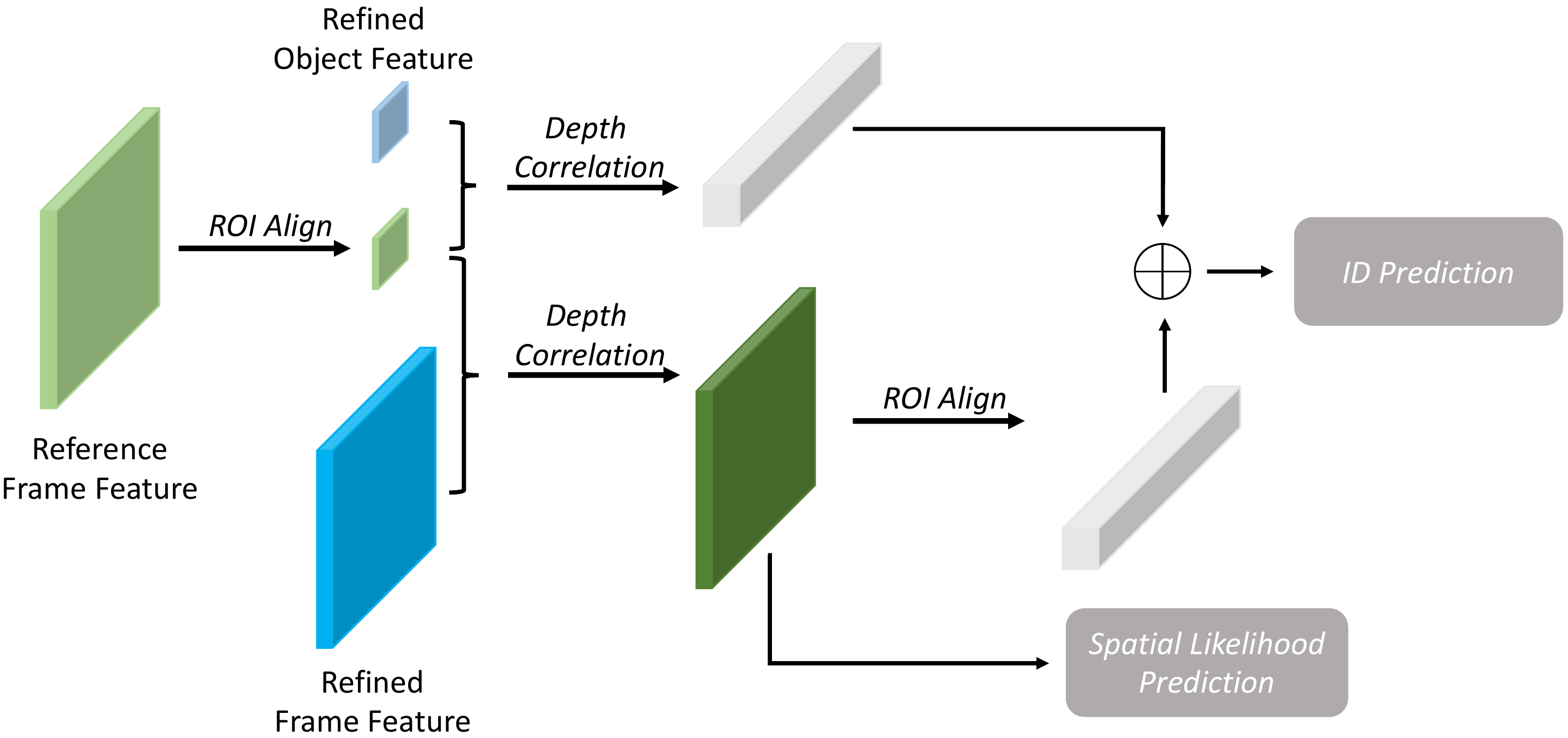}
% 	\vspace{3mm}
	\caption{An illustration of the correlation-based tracking module. The convolution layers are omitted here.}
	\label{fig:track}
\end{figure}

In order to track an object across frames in a more robust and consistent way, we propose a new correlation-based tracking module to generate both spatial likelihoods and object similarities, which is more powerful than the previous approach~\cite{yang2019video}.

Recent works on object tracking shows the effectiveness of correlation-based tracking which has achieved state-of-the-art performance on several tracking benchmarks. Inspired by SiamRPN++~\cite{li2019siamrpn++}, our proposed tracking module incorporates knowledge from both object similarities and cross-correlations of the target object and the search region. As shown in Fig~\ref{fig:track}, our tracking module can be abstracted into three procedures:(a) pair-wise similarity computation, which uses depth-wise correlation instead of matrix multiplication~\cite{yang2019video} to predict a similarity matrix between two groups of objects. (b) cross-correlation computation, which generates correlation maps with depth-wise correlation between a single object and the whole frame. To train the model to produce high quality correlation maps, we employ a pseudo likelihood map represented by a two dimensional gaussian distribution centered at the location of the object as the supervision signal. Then we compute the aggregated similarity vector for a detected object with ROI align within the corresponding bounding box on the correlation map. 
After that, we sum up the two similarity vectors to obtain the final depth-wise similarity vector. (c) The final similarity score is mapped from the similarity vector with  two $1 \times 1$ convolution layers. The proposed correlation-based tracking module considers both feature similarity and correlation-based similarity scores on the raw image features to increase the robustness of tracking.

%------------------------------------------------------------------------
\section{Experiments}\label{exp}
In this section, we conduct extensive experiments on YouTube-VIS~\cite{yang2019video} to evaluate the effectiveness of each proposed component and compare our proposed method with previous approaches.
\subsection{Data and Evaluation Metric}
{\bf Data} YouTube-VIS is the first and largest dataset for video instance segmentation, which is a subset of YouTube-VOS dataset~\cite{xu2018youtube}.
% , which is the large-scale video segmentation dataset. 
YouTube-VIS is comprised of 2,883 high resolution YouTube videos with 40 common object categories. In each video, several objects with bounding boxes and masks are labeled manually and the identities cross different frames are annotated as well. Since only the validation set is available for evaluation, all results reported in this paper are evaluated on the validation set.

{\bf Evaluation Metrics}
To evaluate the performance of the proposed method, we use the metrics mentioned in~\cite{yang2019video}, which are average precision(AP) and average recall(AR) based on a spatial-temporal Intersection-over-Union (IoU). Following the COCO evaluation, AP is computed by averaging over multiple IoU thresholds, \eg from from 50\% to 95\% at step 5\% and AR is the maximum recall given some fixed number of segmented instances per video. Both metrics are first calculated for each category and then averaged over 40 categories. 

\subsection{Implementation Details}

{\bf Training}. Our proposed method is built on Mask-RCNN~\cite{he2017mask}. The backbone network structure is ResNet50 with FPN~\cite{lin2017feature}, which is pre-trained on MSCOCO dataset~\cite{lin2014microsoft}. 
% Then, we adopt the proposed temporal and spatial attention module to the feature maps after the last convolution layer before FPN. For attention module on the object level, the two input features are the region features after ROI Align and the whole frame feature after the second last stage of FPN. 
In the tracking branch, we use two convolution layers to refine the correlation features generated by depth-wise correlations, respectively. The first convolution layer has 256 channels which is the same dimension as the correlation features while the second one is used for correlation map prediction with only one output channel. 
% We use the same structure to predict the correlation score. 
Our model is implemented based on MMDetection~\cite{chen2019mmdetection} and the whole framework is trained end-to-end in 12 epochs with two NVIDIA 2080TI GPUs. We resize the original frame size to $640 \times 360$ for both training and testing. During training, the initial learning rate is set to 0.0125 and decays with a factor of 10 at epoch 8 and 11. For each input frame, we randomly select three frames from the same video, two used as support frames in the dual attention module and the other used as reference frame in the tracking module.

{\bf Testing}. During evaluation, the testing video is processed by the proposed method frame by frame in an online fashion. For each input frame, four additional frames are sampled from the testing video as support frames. Note that the number of support frames used in training and testing can be different, since testing with more support frames can help improve performance. We conduct an ablation study on the number of support frames in the following section. For the tracking head, we ignore the correlation map and use the predicted correlation score as the tracking score. Then we follow the inference procedure described in~\cite{yang2019video} to predict the category, bounding box and mask for the object instance. In addition, we combine other cues, \ie detection confidence, bounding box IoU, and category consistency, along with tracking scores to improve the tracking accuracy as a powerful post-processing.

\subsection{Ablation Study}
\begin{table}\setlength{\tabcolsep}{1pt}
\centering
\footnotesize
\begin{tabular} {l|c|c|c}
\hlineB{2}
Methods & AP & AP$_{0.5}$ & AP$_{0.75}$ \\ \hline 
Mask-Track RCNN~\cite{yang2019video} & 21.1 & 37.7 & 23.6 \\ \hline
Our Implementation & 20.9 & 37.9 & 21.6 \\ 
Our Implementation + MSCOCO & {\bf 24.1} & {\bf 42.6} & {\bf 24.9} \\ \hline
\hlineB{2}
\end{tabular}
% \vspace{1mm}
\caption{Performance of the baseline model on YouTube-VIS validation set. ``Our Implementation" means our reproduced results of Mask-Track RCNN.``Our Implementation + MSCOCO" is used as the baseline model in all ablation studies hereafter.}
\label{exp:t1}
\end{table}

\begin{table}\setlength{\tabcolsep}{4pt}
\centering
\footnotesize
\begin{tabular} {l|c|c|c}
\hlineB{2}
% \multicolumn{4}{c}{(a) {\bf Baseline}}\\ 
% & AP & AP$_{0.5}$ & AP$_{0.75}$ \\ \hline 
% Baseline & 24.1 & 42.6 & 24.9 \\ \hline 
\multicolumn{4}{c}{(a) {\bf Attention on Frame Level} }\\ 
& AP & AP$_{0.5}$ & AP$_{0.75}$ \\ \hline 
Baseline + Temporal Attention & 25.2 & 43.9 & 25.6 \\
Baseline + Spatial Attention & 24.9 & 43.4 & 25.2 \\
Baseline + Spatial-Temporal Attention & 25.8 & 44.5 & 27.0 \\ \hline
\multicolumn{4}{c}{(b) {\bf Attention on Object Level}}\\ 
& AP & AP$_{0.5}$ & AP$_{0.75}$ \\ \hline 
Baseline + Temporal Attention  & 25.4 & 44.4 & 25.9  \\ 
Baseline + Spatial Attention  & 24.8 & 43.3 & 25.1 \\ 
Baseline + Spatial-Temporal Attention & 26.1 & 45.6	& 26.7 \\ \hline
\multicolumn{4}{c}{(c) {\bf Attention on Both Frame and  Object Level}}\\ 
& AP & AP$_{0.5}$ & AP$_{0.75}$ \\ \hline 
Baseline + Spatial-Temporal Attention & {\bf 27.5} & {\bf 46.1} & {\bf 28.9} \\
\hlineB{2}
\end{tabular}

\caption{Ablation Study of our proposed attention module on YouTube-VIS validation set. The best results are highlighted in bold.}
\label{exp:t2}
\end{table}
{\bf Baseline Model and Data Augmentation}
Since our proposed method is built on Mask-Track RCNN~\cite{yang2019video}, we take Mask-Track RCNN as the baseline model to validate the effectiveness of each contribution. We first reproduce the Mask-Track RCNN with publicly available codes and the results are listed in Table~\ref{exp:t1}. Our result is close to the result reported in original paper. Note that all results in Table~\ref{exp:t1} are without any post-processing. 

In addition, we find that the number of object instances in YouTube-VIS dataset is limited. Since the proposed method does not depend on the temporal smoothness of a video, some image-based datasets can be adopted to increase the training samples. We choose MSCOCO~\cite{lin2014microsoft} as external data which has a large overlap on the object categories with YouTube-VIS. In order to make use of the image data, we generate support frames and reference frame based on a single image by some affine transformations, \ie \ie rotation, translation and shearing.
% . Specifically, we first sample the images which only have the objects belonging to categories in YouTube-VIS and then simulate video-style data using data augmentations, \ie rotation, translation and shearing. 
The identity annotations across different images can be generated automatically in this process. The performance after using external data is listed in Table~\ref{exp:t1} as well.
% By using external data from MSCOCO, the performance improves by more than 3\% on AP, AP$_{0.5}$, and AP$_{0.75}$, as reported in Table~\ref{exp:t1}.
We use this model as a baseline model for all the following ablation experiments.

\begin{table}\setlength{\tabcolsep}{2pt}
\centering
\footnotesize
\begin{tabular} {l|c|c|c}
\hlineB{2}
% \multicolumn{4}{c}{(a) {\bf Baseline}}\\
% & AP & AP$_{0.5}$ & AP$_{0.75}$ \\ \hline 
% Baseline & 24.1 & 42.6 & 24.9 \\ \hline 
% \multicolumn{4}{c}{{\bf Correlation Map}} \\ \hline
Modules & AP & AP$_{0.5}$ & AP$_{0.75}$ \\ \hline 
Baseline + CM & 25.1 & 44.3 & 26.7 \\
Baseline + CM + FDA & 26.3 & 45.3 & 27.1 \\
Baseline + CM + BDA & 26.7 & 45.9 & 27.3 \\
Baseline + CM + FDA + BDA (CompFeat) & {\bf 28.4} & {\bf 47.4} & {\bf 30.7} \\
\hlineB{2}
\end{tabular}
% \vspace{1mm}
\caption{Ablation Study of the proposed track module on YouTube-VIS validation set. CM, FDA, BDA denote the proposed tracking module with correlation map, the frame level dual attention module and object level dual attention module, respectively. The best results are highlighted in bold.}
\label{exp:t3}
\end{table}

% \begin{table}\setlength{\tabcolsep}{10pt}
% \centering
% \footnotesize
% \begin{tabular} {l|c|c}
% \hlineB{2}
% Methods & Time (ms) & AP \\ \hline 
% Baseline & 34 & 24.1(32.2) \\
% CompFeat (2 frames) & 61 & 27.3(34.4) \\ 
% CompFeat (3 frames) & 67 & 27.7(34.7) \\
% CompFeat (4 frames) & 78 & 28.4(35.3) \\ 
% CompFeat (5 frames) & 89 & 27.4(34.5)\\\hline
% \hlineB{2}
% \end{tabular}
% \vspace{1mm}
% \caption{Running time analysis on the validation set of Youtube-VIS. The running time is reported in microseconds. The numbers in brackets are the running time including post-processing.}
% \label{exp:t5}
% \end{table}

\begin{table}\setlength{\tabcolsep}{3pt}
\centering
\footnotesize
\begin{tabular} {l|c|c|c|c}
\hlineB{2}
Train/Test(Uniform) &  2 frames &  3 frames &   3 frames & 4 frames \\ \hline 
Uniform 2 frames & 27.1 & 27.4 & 27.8 & 27.3  \\
Uniform 3 frames & 26.2 & 26.6 & 26.6 & 26.8  \\ 
Random 2 frames & 27.3 & 27.7 & 28.4 & 27.4\\
Random 3 frames  & 26.1 & 26.8 & 26.8 & 26.7 \\ \hline
\hlineB{2}
\end{tabular}
% \vspace{1mm}
\caption{Performance of different sampling methods and different frames during training/testing on the validation set of Youtube-VIS. The performance is reported in AP.}
\label{exp:t6}
\end{table}

\begin{table*}[t]\setlength{\tabcolsep}{6pt}
\centering
\footnotesize
\begin{tabular} {c|c|c|c|c|c}
\hlineB{2}
Methods & AP & AP$_{0.5}$ & AP$_{0.75}$ & AR$_1$ & AR$_{10}$ \\ \hline
% OSMN~\cite{yang2018efficient}  & 23.4 & 36.5 & 25.7 & 28.9 & 31.1 \\
% FEELVOS~\cite{voigtlaender2019feelvos} & 26.9 & 42.0 & 29.7 & 29.9 & 33.4 \\
IoUTracker+~\cite{bochinski2017high}  & 23.6 & 39.2 & 25.5 & 26.2 & 30.9 \\
OSMN~\cite{yang2018efficient} & 27.5 & 45.1 & 29.1 & 28.6 & 33.1 \\
DeepSORT~\cite{wojke2017simple} & 26.1 & 42.9& 26.1 & 27.8 & 31.3 \\
SeqTracker & 27.5 & 45.7 & 28.7 & 29.7 & 32.5 \\
% MaskTrack R-CNN$^*$~\cite{yang2019video} & 21.1 & 37.7 & 23.6 & -- &-- \\
MaskTrack R-CNN~\cite{yang2019video} & 30.3 & 51.1 & 32.6 & 31.0 & 35.5 \\
SipMask~\cite{Cao_SipMask_ECCV_2020} & 32.5 & 53.0 & 33.3 & 33.5 & 38.9 \\
\hline
MaskTrack R-CNN + MSCOCO & 32.2 & 52.1 & 34.6 & 31.8 & 37.2 \\
CompFeat & {\bf 35.3} & {\bf 56.0} & {\bf 38.6} & {\bf 33.1} & {\bf 40.3} \\
\hlineB{2}
\end{tabular}
% \vspace{1mm}
\caption{Comparison of the proposed approach with the state-of-the-arts on YouTube-VIS validation set. Note that all results in this Table including the post-processing. The best results are highlighted in bold.}
\label{exp:t4}
\end{table*}

{\bf Effectiveness of Attention Module}
We conduct ablation study to prove the effectiveness of the proposed attention module: temporal attention and spatial attention. The results are listed in Table~\ref{exp:t2}. Note all results here are without post-processing. We first evaluate each attention module on frame level in Table~\ref{exp:t2}(a). With the temporal attention module, we improve baseline model by 1.1\% in AP and  1.3\% in AP$_{0.5}$ respectively. Similarly, spatial attention module can also slightly improve the baseline performance by 1\%. When combining both temporal and spatial attention together as the dual attention module, we further boost the baseline model by 1.7\%, 1.9\% and 2.1\% in AP, AP$_{0.5}$ and AP$_{0.75}$. These experimental results prove that our proposed attention module can aggregate helpful context feature from both other frames and the input frame. 

We then experiment with temporal and spatial attention on object level. Table~\ref{exp:t2}(b) shows that proposed attention method on object level can always achieve comparable or even better results compared with the frame level one. For instance, by performing both temporal and spatial attention on object level, the performance gain becomes 2.0\%, 3.0\% and 1.8\% in AP, AP$_{0.5}$ and AP$_{0.75}$ respectively. 
% Note that we only adopt the object level attention module on the detection branch, because we obtain slightly worse performance when apply it to both detection and mask generation branch.

Furthermore, Table~\ref{exp:t2}(c) lists the performance of combining attention modules on both frame level and object level. Comparing with the performance only on frame or object level, the combination one is superior. Specifically, with both attention on two different levels, we achieve AP/AP$_{0.5}$ $=27.5\%/46.1\%$, which outperforms the performance with attention module on frame level by 1.7\%/1.6\%. 
% The similar improvement can also be found when compared with attention module on object level. 
This further improvement shows that by using attention module on frame level and object level, we can aggregate context information in a global-to-local manner, which can greatly improve the baseline model by 3.4\%, 3.5\% and 4.0\% in AP, AP$_{0.5}$ and AP$_{0.75}$.

\begin{figure}[t]
\begin{center}
\bgroup 
 \def\arraystretch{0.1} 
 \setlength\tabcolsep{0.5pt}
\begin{tabular}{ccccc}
\includegraphics[width=0.2\linewidth]{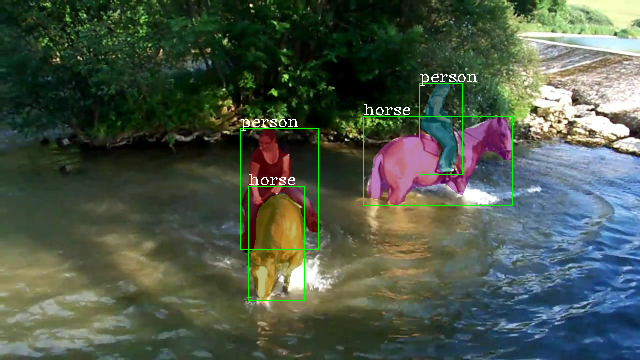} &
\includegraphics[width=0.2\linewidth]{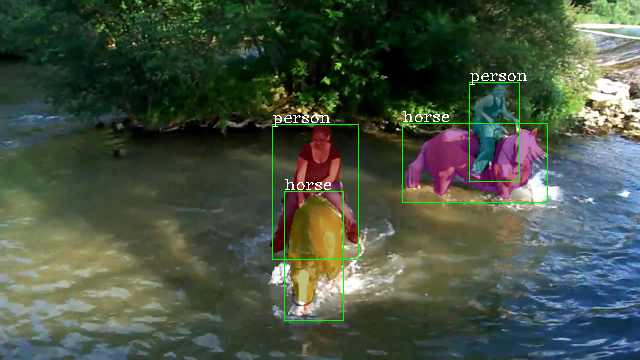} & 
\includegraphics[width=0.2\linewidth]{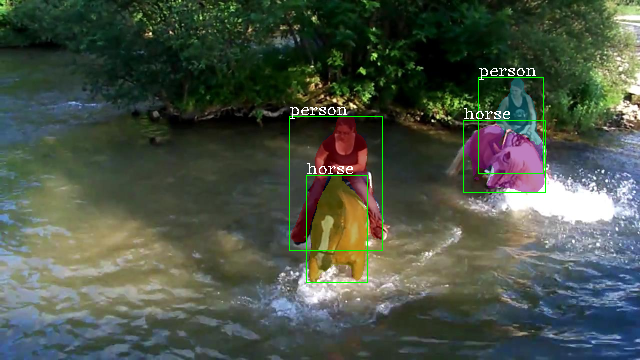} & 
\includegraphics[width=0.2\linewidth]{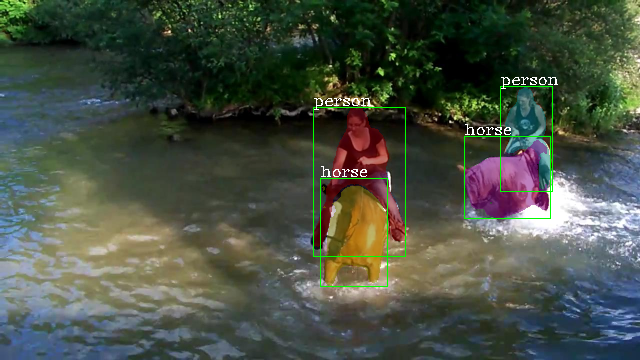} &
\includegraphics[width=0.2\linewidth]{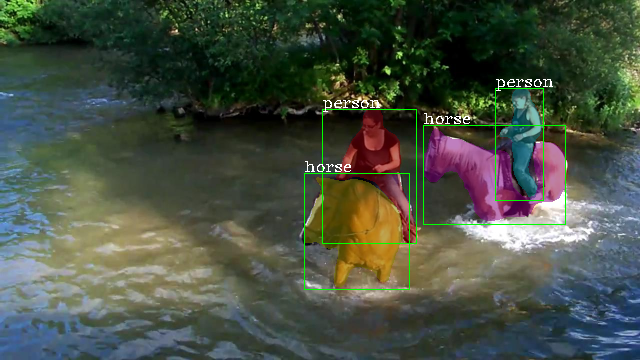} \\
\includegraphics[width=0.2\linewidth]{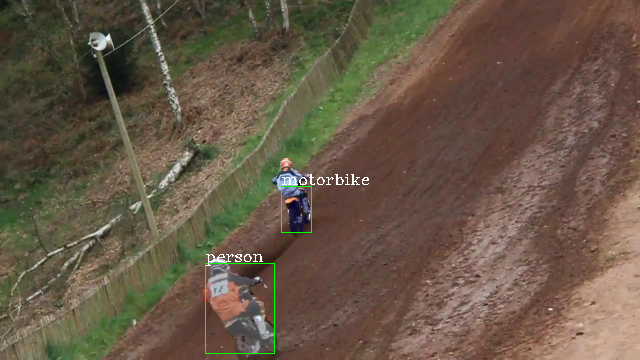} &
\includegraphics[width=0.2\linewidth]{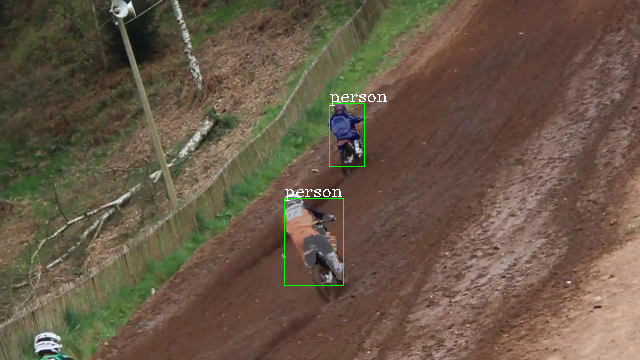} &
\includegraphics[width=0.2\linewidth]{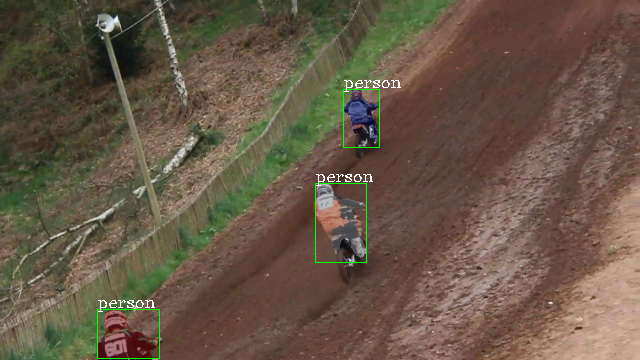} &
\includegraphics[width=0.2\linewidth]{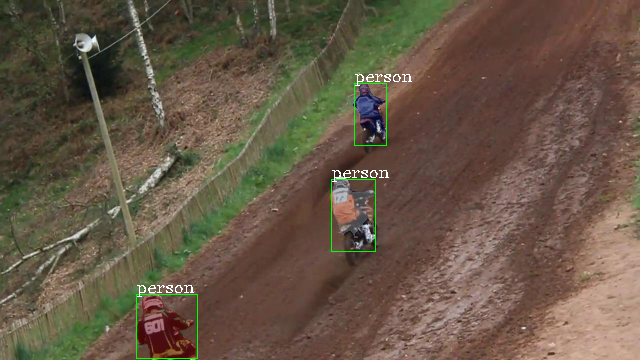} &
\includegraphics[width=0.2\linewidth]{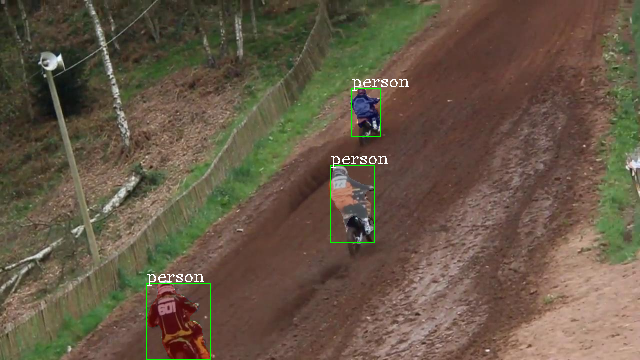} \\
\includegraphics[width=0.2\linewidth]{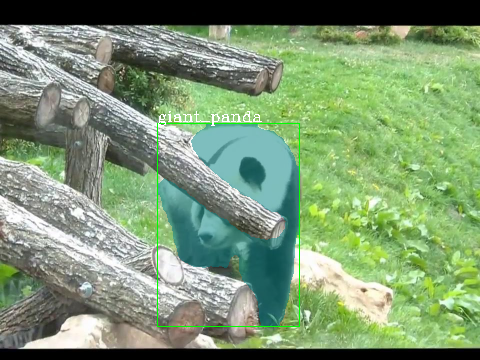} &
\includegraphics[width=0.2\linewidth]{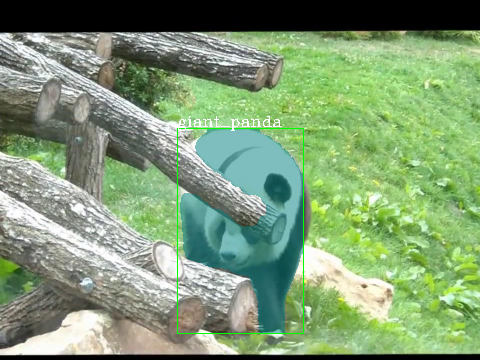} &
\includegraphics[width=0.2\linewidth]{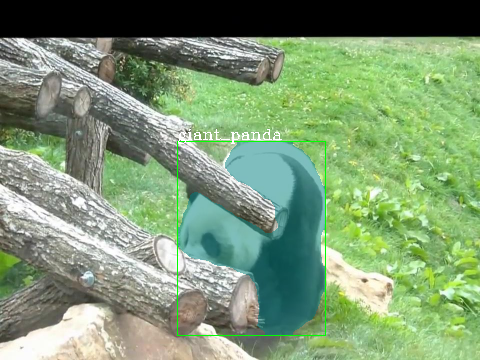} &
\includegraphics[width=0.2\linewidth]{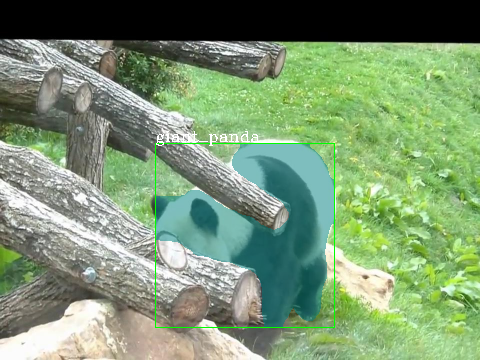} &
\includegraphics[width=0.2\linewidth]{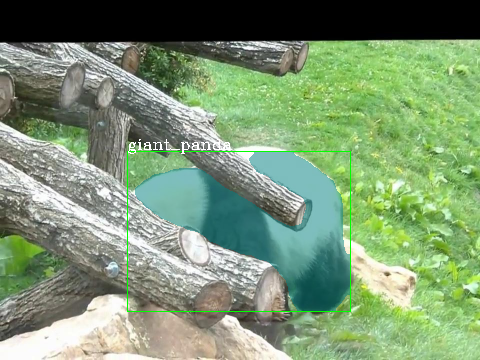} \\
\includegraphics[width=0.2\linewidth]{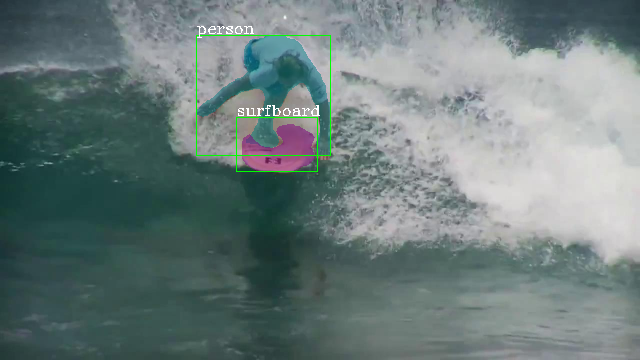} &
\includegraphics[width=0.2\linewidth]{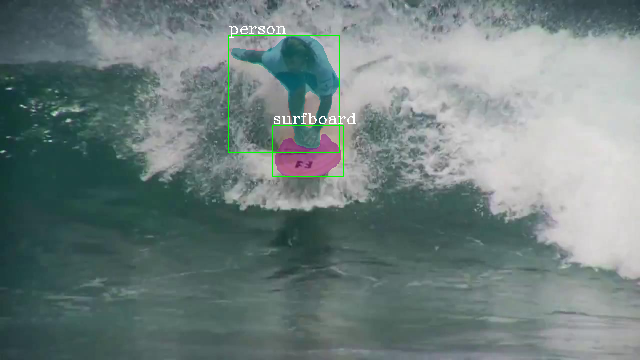} &
\includegraphics[width=0.2\linewidth]{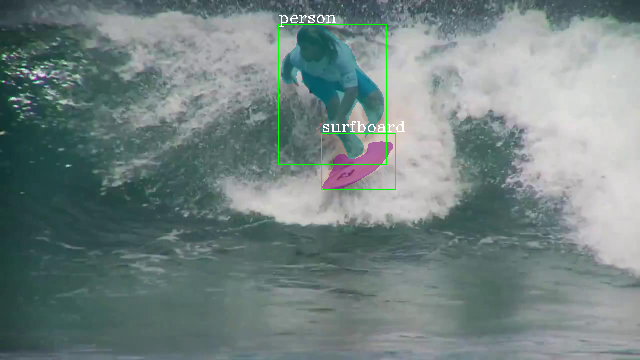} &
\includegraphics[width=0.2\linewidth]{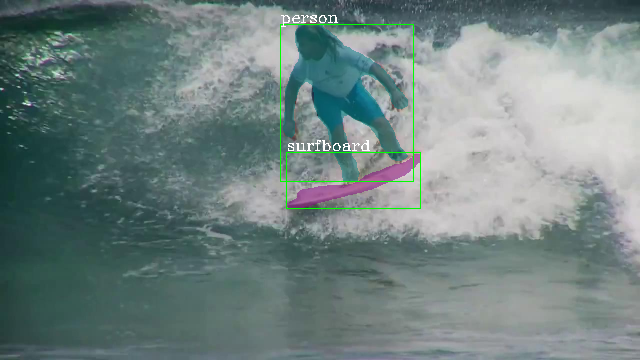} &
\includegraphics[width=0.2\linewidth]{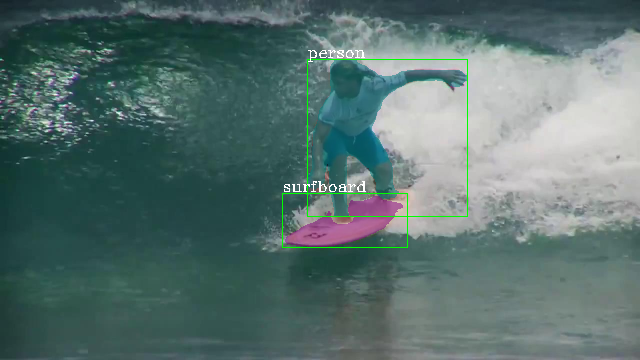} \\
\includegraphics[width=0.2\linewidth]{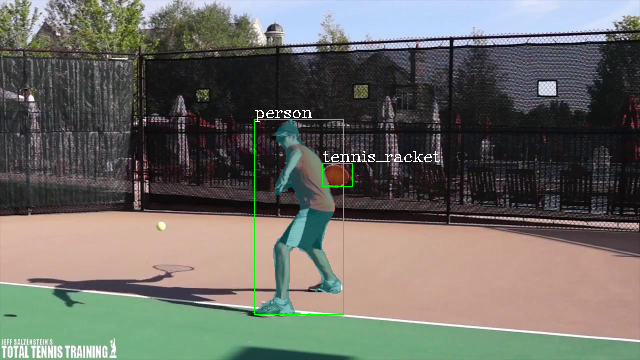} &
\includegraphics[width=0.2\linewidth]{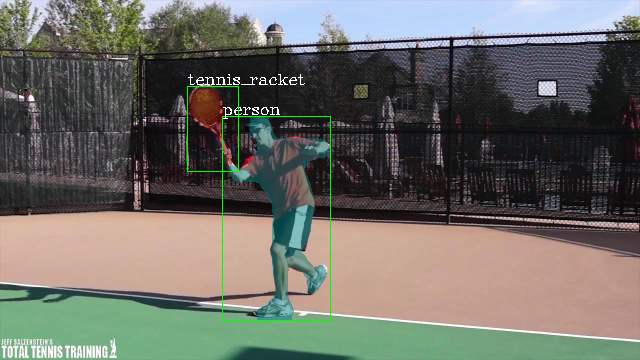} &
\includegraphics[width=0.2\linewidth]{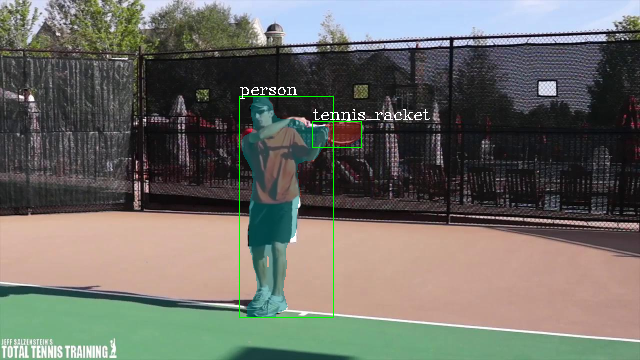} &
\includegraphics[width=0.2\linewidth]{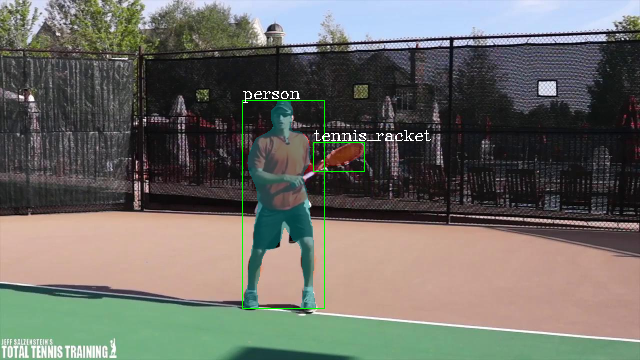} &
\includegraphics[width=0.2\linewidth]{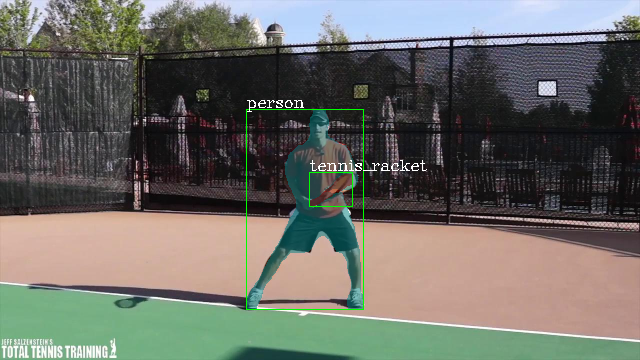} \\
\includegraphics[width=0.2\linewidth]{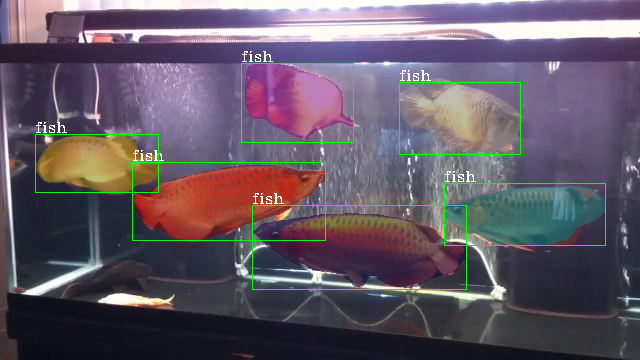} &
\includegraphics[width=0.2\linewidth]{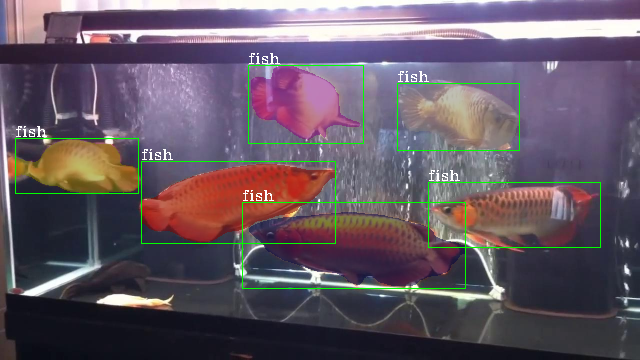} &
\includegraphics[width=0.2\linewidth]{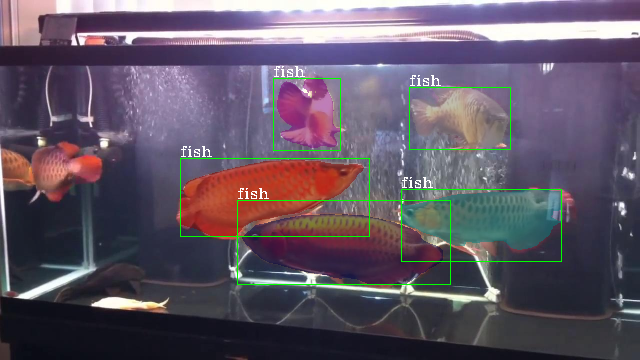} &
\includegraphics[width=0.2\linewidth]{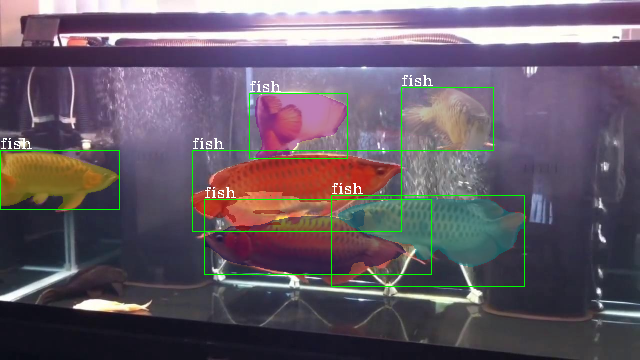} &
\includegraphics[width=0.2\linewidth]{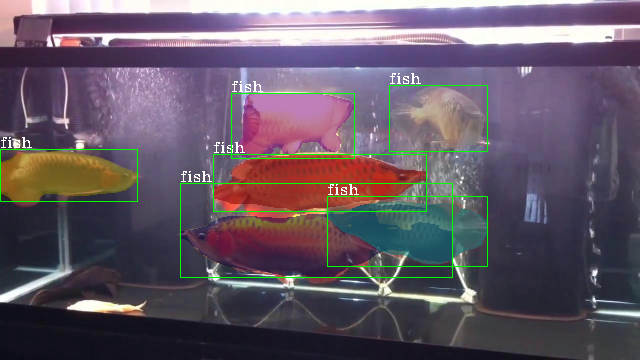} \\
\end{tabular} \egroup
\end{center}
\caption{Visualization results of CompFeat. Each row has five sampled frames from a video sequence. Categories, bounding boxes and instance masks are shown for each object. Note objects with the same predicated identity across frames are marked with the same color. Zoom in to see details.}
\label{exp:v1}
\end{figure}

{\bf Effectiveness of Correlation-based Tracking Module}
The ablation study on the proposed correlation-based tracking module is shown in Table~\ref{exp:t3}. Again, the results are without post-processing. Compared to the baseline model, the tracking module with correlation map outperforms the baseline model by more than 1\% on AP, AP$_{0.5}$ and AP$_{0.75}$. This improvement indicates the cross-correlation map between objects and the whole frames contains more semantic information than the object features used by the baseline. In addition, when integrating it into the whole framework with dual attention on frame level or object level, we obtain consistently better performance. In particular, by using correlation maps and dual attention module on frame level, we improve the performance in AP from 25.8\% to 26.3\%. Finally, we evaluate the performance of video instance segmentation with all our proposed modules, \eg frame level dual attention module, object level dual attention module and correlation maps. As shown in the last row in Table~\ref{exp:t3}, we obtain the best performance. For instance, we achieve 28.4\% and 47.4\% on AP and $AP_{0.5}$ with all proposed modules, which surpasses the baseline model by 4.3\% and 4.8\%. 

% {\bf Running Time Analysis.} The running time analysis is listed in Table~\ref{exp:t5}. The runing time is tested with a single NVIDIA 2080TI GPU. With more support frames used in CompFeat module, we can gradually improve its performance but the running time increases as well. Our best performance is achieved with four support frames where each frame is processed in 78ms and AP is 28.4\%. A potential reason for the performance degeneration with more frames is the training-testing mismatch of using support frames.

{\bf Sampling Method.} The VIS performance under different sampling methods are shown in Table~\ref{exp:t6}. Note that, for fairness, all experiment use the uniform sampling during testing. From Table~\ref{exp:t6}, it can be observed that the random sampling is always powerful than the uniform one during training. The reason is that random sampling increases the variety of samples during training, which can lead our proposed ComFeat model discovers more temporal and spatial correspondences between current frames and support frames.

% In conclusion, we evaluate each proposed module by several experiments and the best model can be achieved when combining all of them together into an end-to-end framework. Next, we compare the best model with previous approaches.

\subsection{Comparison with the State-of-the-Arts}
Mask-Track RCNN~\cite{yang2019video} is the first work on video instance segmentation. There are several work proposed in the Large-Scale Video Object Segmentation challenge~\cite{luiten2019video,wang2019empirical, dong2019temporal}, but it is hard to compare with them since they use different backbone networks and different external training data. We borrow experimental results of other approaches from~\cite{yang2019video}. The comparison results are presented in Table~\ref{exp:t4}. 
% IoUTracker+~\cite{bochinski2017high} assigns the instance label with the largest score to a candidate box. Since it does not leverage any visual information, its performance is a little weaker. OSMN~\cite{yang2018efficient} estimates a new mask of the instance at a new frame using VOS algorithm and is more robust to occlusion and fast motion. DeepSORT~\cite{wojke2017simple} measures the appearance similarity between bounding boxes with a deep neural network and achieves track matching with both IOU score and visual similarity score, which achieves better performance the IoUTracker+. SeqTracker first computes instance segmentation results for all frames of a video, and then searches all possible tracks to find the one with the largest score. It is an offline method which cannot predict instance tracks sequentially. 
Mask-Track RCNN~\cite{yang2019video} is an online method which learns feature similarities for object matching. And SipMask~\cite{Cao_SipMask_ECCV_2020} shares the similar structure while replace the instance segmentation branch with an one stage instance segmentation module. Compared with these methods, our proposed CompFeat achieves the best performance under all evaluation metrics. Compared with our baseline Mask-Track RCNN, the proposed CompFeat outperform it by a large margin. 
% For instance, the performance on AP, AP$_{0.5}$ and AP$_{0.75}$ is improved by 5.0\%, 4.9\% and 6.0\% respectively. 
Note this performance gain is not from the additional training data since MaskTrack R-CNN with the same training data only achieves 32.2\% and 52.1\% on AP and AP$_{0.5}$.

\subsection{Qualitative Results}
Fig.~\ref{exp:v1} shows some qualitative results of our proposed CompFeat on YouTube-VIS validation set. Each row represents the predicted results on different frames in a video. The objects with the same identity are shown in the same color. As shown, CompFeat makes accurate predictions on object categories, bounding boxes, masks and identities under challenging conditions, \ie  multiple similar objects (row 1, 2), moderate occlusions (row 3), and drastic appearance changes (row 4).
% Our method successfully generates video instance segmentations under multiple similar objects (row 1, 2), moderate occlusions (row 3), and drastic appearance changes (row 4).
The last row shows a challenging case with six fish where our algorithm performs much better than MaskTrack-RCNN~\cite{yang2019video} although it misses a fish in the third image. 
% Last row shows a failure case. Our algorithm mistakenly groups two apes into one in several intermediate frames and switches the identities of them at the end of the video. 

%------------------------------------------------------------------------
% \section{Discussions}
% \input{TEX/5_discussion.tex}
%------------------------------------------------------------------------
\section{Conclusion}
In this paper, we develop a comprehensive approach for feature aggregation for video instance segmentation, which is an underexplored direction in this area.
Attention mechanisms are careful crafted for feature aggregations on both frame-level and object-level in both temporal and spatial manner.
A new tracking module is  designed to enhance local discriminative power of features with local and global correlation maps, in order to improve robustness of object tracking and re-identification. The effectiveness of the proposed modules is systematically evaluated with extensive experiments and ablation studies on the YouTube-VIS dataset.
% We believe the proposed feature aggregation approach could be beneficial to other video recognition tasks such as video object detection and video action detection. 
%------------------------------------------------------------------------

\bibliography{sample-base}

\begin{thebibliography}{38}
\providecommand{\natexlab}[1]{#1}
\providecommand{\url}[1]{\texttt{#1}}
\providecommand{\urlprefix}{URL }
\expandafter\ifx\csname urlstyle\endcsname\relax
  \providecommand{\doi}[1]{doi:\discretionary{}{}{}#1}\else
  \providecommand{\doi}{doi:\discretionary{}{}{}\begingroup
  \urlstyle{rm}\Url}\fi

\bibitem[{Bochinski, Eiselein, and Sikora(2017)}]{bochinski2017high}
Bochinski, E.; Eiselein, V.; and Sikora, T. 2017.
\newblock High-speed tracking-by-detection without using image information.
\newblock In \emph{2017 14th IEEE International Conference on Advanced Video
  and Signal Based Surveillance (AVSS)}, 1--6. IEEE.

\bibitem[{Cao et~al.(2020)Cao, Anwer, Cholakkal, Khan, Pang, and
  Shao}]{Cao_SipMask_ECCV_2020}
Cao, J.; Anwer, R.~M.; Cholakkal, H.; Khan, F.~S.; Pang, Y.; and Shao, L. 2020.
\newblock SipMask: Spatial Information Preservation for Fast Instance
  Segmentation.
\newblock \emph{Proc. European Conference on Computer Vision} .

\bibitem[{Cao et~al.(2019)Cao, Xu, Lin, Wei, and Hu}]{cao2019gcnet}
Cao, Y.; Xu, J.; Lin, S.; Wei, F.; and Hu, H. 2019.
\newblock GCNet: Non-local Networks Meet Squeeze-Excitation Networks and
  Beyond.
\newblock \emph{arXiv preprint arXiv:1904.11492} .

\bibitem[{Chen et~al.(2019)Chen, Wang, Pang, Cao, Xiong, Li, Sun, Feng, Liu, Xu
  et~al.}]{chen2019mmdetection}
Chen, K.; Wang, J.; Pang, J.; Cao, Y.; Xiong, Y.; Li, X.; Sun, S.; Feng, W.;
  Liu, Z.; Xu, J.; et~al. 2019.
\newblock MMDetection: Open MMLab Detection Toolbox and Benchmark.
\newblock \emph{arXiv preprint arXiv:1906.07155} .

\bibitem[{Chen et~al.(2018)Chen, Wang, Yang, Zhang, Xiong, Change~Loy, and
  Lin}]{chen2018optimizing}
Chen, K.; Wang, J.; Yang, S.; Zhang, X.; Xiong, Y.; Change~Loy, C.; and Lin, D.
  2018.
\newblock Optimizing video object detection via a scale-time lattice.
\newblock In \emph{Proceedings of the IEEE Conference on Computer Vision and
  Pattern Recognition}, 7814--7823.

\bibitem[{Dong et~al.(2019)Dong, Wang, Huang, Yu, Su, Zhou, Shao, Wen, and
  Wang}]{dong2019temporal}
Dong, M.; Wang, J.; Huang, Y.; Yu, D.; Su, K.; Zhou, K.; Shao, J.; Wen, S.; and
  Wang, C. 2019.
\newblock Temporal Feature Augmented Network for Video Instance Segmentation.
\newblock In \emph{Proceedings of the IEEE International Conference on Computer
  Vision Workshops}.

\bibitem[{Feichtenhofer, Pinz, and Zisserman(2017)}]{feichtenhofer2017detect}
Feichtenhofer, C.; Pinz, A.; and Zisserman, A. 2017.
\newblock Detect to track and track to detect.
\newblock In \emph{{IEEE ICCV}}.

\bibitem[{Han et~al.(2016)Han, Khorrami, Paine, Ramachandran, Babaeizadeh, Shi,
  Li, Yan, and Huang}]{han2016seq}
Han, W.; Khorrami, P.; Paine, T.~L.; Ramachandran, P.; Babaeizadeh, M.; Shi,
  H.; Li, J.; Yan, S.; and Huang, T.~S. 2016.
\newblock Seq-nms for video object detection.
\newblock \emph{arXiv preprint arXiv:1602.08465} .

\bibitem[{Hariharan et~al.(2014)Hariharan, Arbel{\'a}ez, Girshick, and
  Malik}]{hariharan2014simultaneous}
Hariharan, B.; Arbel{\'a}ez, P.; Girshick, R.; and Malik, J. 2014.
\newblock Simultaneous detection and segmentation.
\newblock In \emph{{ECCV}}.

\bibitem[{He et~al.(2017)He, Gkioxari, Doll{\'a}r, and Girshick}]{he2017mask}
He, K.; Gkioxari, G.; Doll{\'a}r, P.; and Girshick, R. 2017.
\newblock Mask r-cnn.
\newblock In \emph{{IEEE ICCV}}.

\bibitem[{He et~al.(2016)He, Zhang, Ren, and Sun}]{he2016deep}
He, K.; Zhang, X.; Ren, S.; and Sun, J. 2016.
\newblock Deep residual learning for image recognition.
\newblock In \emph{{IEEE CVPR}}, 770--778.

\bibitem[{Li et~al.(2019)Li, Wu, Wang, Zhang, Xing, and Yan}]{li2019siamrpn++}
Li, B.; Wu, W.; Wang, Q.; Zhang, F.; Xing, J.; and Yan, J. 2019.
\newblock Siamrpn++: Evolution of siamese visual tracking with very deep
  networks.
\newblock In \emph{{IEEE CVPR}}.

\bibitem[{Lin et~al.(2017)Lin, Doll{\'a}r, Girshick, He, Hariharan, and
  Belongie}]{lin2017feature}
Lin, T.-Y.; Doll{\'a}r, P.; Girshick, R.; He, K.; Hariharan, B.; and Belongie,
  S. 2017.
\newblock Feature pyramid networks for object detection.
\newblock In \emph{{IEEE CVPR}}.

\bibitem[{Lin et~al.(2014)Lin, Maire, Belongie, Hays, Perona, Ramanan,
  Doll{\'a}r, and Zitnick}]{lin2014microsoft}
Lin, T.-Y.; Maire, M.; Belongie, S.; Hays, J.; Perona, P.; Ramanan, D.;
  Doll{\'a}r, P.; and Zitnick, C.~L. 2014.
\newblock Microsoft coco: Common objects in context.
\newblock In \emph{{ECCV}}.

\bibitem[{Liu et~al.(2019)Liu, Zhu, White, Li, and
  Kalenichenko}]{liu2019looking}
Liu, M.; Zhu, M.; White, M.; Li, Y.; and Kalenichenko, D. 2019.
\newblock Looking Fast and Slow: Memory-Guided Mobile Video Object Detection.
\newblock \emph{arXiv preprint arXiv:1903.10172} .

\bibitem[{Luiten, Torr, and Leibe(2019)}]{luiten2019video}
Luiten, J.; Torr, P.; and Leibe, B. 2019.
\newblock Video Instance Segmentation 2019: A winning approach for combined
  Detection, Segmentation, Classification and Tracking.
\newblock In \emph{Proceedings of the IEEE International Conference on Computer
  Vision Workshops}.

\bibitem[{Oh et~al.(2019)Oh, Lee, Xu, and Kim}]{oh2019video}
Oh, S.~W.; Lee, J.-Y.; Xu, N.; and Kim, S.~J. 2019.
\newblock Video object segmentation using space-time memory networks.
\newblock \emph{arXiv preprint arXiv:1904.00607} .

\bibitem[{Russakovsky et~al.(2015)Russakovsky, Deng, Su, Krause, Satheesh, Ma,
  Huang, Karpathy, Khosla, Bernstein et~al.}]{russakovsky2015imagenet}
Russakovsky, O.; Deng, J.; Su, H.; Krause, J.; Satheesh, S.; Ma, S.; Huang, Z.;
  Karpathy, A.; Khosla, A.; Bernstein, M.; et~al. 2015.
\newblock Imagenet large scale visual recognition challenge.
\newblock \emph{International journal of computer vision} .

\bibitem[{Sadeghian, Alahi, and Savarese(2017)}]{sadeghian2017tracking}
Sadeghian, A.; Alahi, A.; and Savarese, S. 2017.
\newblock Tracking the untrackable: Learning to track multiple cues with
  long-term dependencies.
\newblock In \emph{{IEEE ICCV}}.

\bibitem[{Shi(2018)}]{shi2018geometry}
Shi, H. 2018.
\newblock Geometry-aware traffic flow analysis by detection and tracking.
\newblock In \emph{Proceedings of the IEEE Conference on Computer Vision and
  Pattern Recognition Workshops}, 116--120.

\bibitem[{Son et~al.(2017)Son, Baek, Cho, and Han}]{son2017multi}
Son, J.; Baek, M.; Cho, M.; and Han, B. 2017.
\newblock Multi-object tracking with quadruplet convolutional neural networks.
\newblock In \emph{{IEEE CVPR}}.

\bibitem[{Valmadre et~al.(2017)Valmadre, Bertinetto, Henriques, Vedaldi, and
  Torr}]{valmadre2017end}
Valmadre, J.; Bertinetto, L.; Henriques, J.; Vedaldi, A.; and Torr, P.~H. 2017.
\newblock End-to-end representation learning for correlation filter based
  tracking.
\newblock In \emph{{IEEE CVPR}}.

\bibitem[{Voigtlaender et~al.(2019{\natexlab{a}})Voigtlaender, Chai, Schroff,
  Adam, Leibe, and Chen}]{voigtlaender2019feelvos}
Voigtlaender, P.; Chai, Y.; Schroff, F.; Adam, H.; Leibe, B.; and Chen, L.-C.
  2019{\natexlab{a}}.
\newblock Feelvos: Fast end-to-end embedding learning for video object
  segmentation.
\newblock In \emph{{IEEE CVPR}}.

\bibitem[{Voigtlaender et~al.(2019{\natexlab{b}})Voigtlaender, Krause, Osep,
  Luiten, Sekar, Geiger, and Leibe}]{voigtlaender2019mots}
Voigtlaender, P.; Krause, M.; Osep, A.; Luiten, J.; Sekar, B. B.~G.; Geiger,
  A.; and Leibe, B. 2019{\natexlab{b}}.
\newblock MOTS: Multi-object tracking and segmentation.
\newblock In \emph{{IEEE CVPR}}.

\bibitem[{Voigtlaender and Leibe(2017)}]{voigtlaender2017online}
Voigtlaender, P.; and Leibe, B. 2017.
\newblock Online adaptation of convolutional neural networks for video object
  segmentation.
\newblock \emph{arXiv preprint arXiv:1706.09364} .

\bibitem[{Wang et~al.(2019)Wang, He, Yang, Yang, and Torr}]{wang2019empirical}
Wang, Q.; He, Y.; Yang, X.; Yang, Z.; and Torr, P. 2019.
\newblock An Empirical Study of Detection-Based Video Instance Segmentation.
\newblock In \emph{Proceedings of the IEEE International Conference on Computer
  Vision Workshops}.

\bibitem[{Wang et~al.(2018{\natexlab{a}})Wang, Teng, Xing, Gao, Hu, and
  Maybank}]{wang2018learning}
Wang, Q.; Teng, Z.; Xing, J.; Gao, J.; Hu, W.; and Maybank, S.
  2018{\natexlab{a}}.
\newblock Learning attentions: residual attentional siamese network for high
  performance online visual tracking.
\newblock In \emph{{IEEE ICCV}}.

\bibitem[{Wang et~al.(2018{\natexlab{b}})Wang, Girshick, Gupta, and
  He}]{wang2018non}
Wang, X.; Girshick, R.; Gupta, A.; and He, K. 2018{\natexlab{b}}.
\newblock Non-local neural networks.
\newblock In \emph{{IEEE CVPR}}.

\bibitem[{Wojke, Bewley, and Paulus(2017)}]{wojke2017simple}
Wojke, N.; Bewley, A.; and Paulus, D. 2017.
\newblock Simple online and realtime tracking with a deep association metric.
\newblock In \emph{{IEEE International Conference on Image Processing}}.

\bibitem[{Wu et~al.(2019)Wu, Feichtenhofer, Fan, He, Krahenbuhl, and
  Girshick}]{wu2019long}
Wu, C.-Y.; Feichtenhofer, C.; Fan, H.; He, K.; Krahenbuhl, P.; and Girshick, R.
  2019.
\newblock Long-term feature banks for detailed video understanding.
\newblock In \emph{Proceedings of the IEEE Conference on Computer Vision and
  Pattern Recognition}, 284--293.

\bibitem[{Wug~Oh et~al.(2018)Wug~Oh, Lee, Sunkavalli, and
  Joo~Kim}]{wug2018fast}
Wug~Oh, S.; Lee, J.-Y.; Sunkavalli, K.; and Joo~Kim, S. 2018.
\newblock Fast video object segmentation by reference-guided mask propagation.
\newblock In \emph{{IEEE CVPR}}.

\bibitem[{Xu et~al.(2019)Xu, Wen, Li, Bo, and Huang}]{xu2019spatiotemporal}
Xu, K.; Wen, L.; Li, G.; Bo, L.; and Huang, Q. 2019.
\newblock Spatiotemporal CNN for Video Object Segmentation.
\newblock In \emph{{IEEE CVPR}}.

\bibitem[{Xu et~al.(2018)Xu, Yang, Fan, Yue, Liang, Yang, and
  Huang}]{xu2018youtube}
Xu, N.; Yang, L.; Fan, Y.; Yue, D.; Liang, Y.; Yang, J.; and Huang, T. 2018.
\newblock Youtube-vos: A large-scale video object segmentation benchmark.
\newblock \emph{arXiv preprint arXiv:1809.03327} .

\bibitem[{Yang, Fan, and Xu(2019)}]{yang2019video}
Yang, L.; Fan, Y.; and Xu, N. 2019.
\newblock Video Instance Segmentation.
\newblock In \emph{{IEEE ICCV}}.

\bibitem[{Yang et~al.(2018)Yang, Wang, Xiong, Yang, and
  Katsaggelos}]{yang2018efficient}
Yang, L.; Wang, Y.; Xiong, X.; Yang, J.; and Katsaggelos, A.~K. 2018.
\newblock Efficient video object segmentation via network modulation.
\newblock In \emph{{IEEE CVPR}}.

\bibitem[{Zhang and Peng(2019)}]{zhang2019deeper}
Zhang, Z.; and Peng, H. 2019.
\newblock Deeper and wider siamese networks for real-time visual tracking.
\newblock In \emph{{IEEE CVPR}}.

\bibitem[{Zhu et~al.(2017)Zhu, Wang, Dai, Yuan, and Wei}]{zhu2017flow}
Zhu, X.; Wang, Y.; Dai, J.; Yuan, L.; and Wei, Y. 2017.
\newblock Flow-guided feature aggregation for video object detection.
\newblock In \emph{{IEEE ICCV}}.

\bibitem[{Zhu et~al.(2018)Zhu, Wang, Li, Wu, Yan, and Hu}]{Zhu_2018_ECCV}
Zhu, Z.; Wang, Q.; Li, B.; Wu, W.; Yan, J.; and Hu, W. 2018.
\newblock Distractor-aware Siamese Networks for Visual Object Tracking.
\newblock In \emph{{ECCV}}.

\end{thebibliography}

\end{document}